\documentclass{article}

    \PassOptionsToPackage{numbers, compress}{natbib}

    \usepackage[preprint]{neurips_2024}

\usepackage[utf8]{inputenc} 
\usepackage[T1]{fontenc}    
\usepackage{hyperref}       
\usepackage{url}            
\usepackage{booktabs}       
\usepackage{amsfonts}       
\usepackage{nicefrac}       
\usepackage{microtype}      
\usepackage{xcolor}         

\usepackage{amsmath}
\usepackage{graphicx}
\usepackage{algorithmic}
\usepackage{algorithm}
\usepackage{listings}
\usepackage{multicol}
\usepackage{multirow}
\usepackage{makecell}
\usepackage{pifont}
\usepackage{tikz}
\usepackage{caption}

\usepackage{placeins}

\def\eg{\emph{e.g.}}
\def\ie{\emph{i.e.}}

\def\etal{\emph{et al. }}

\newcommand\logicmixor{%
  \mathrel{\ooalign{\hss$\lor$\hss\cr%
  \kern0.4ex\raise0.7ex\hbox{\scalebox{0.7}{$?$}}}}}

\definecolor{codegreen}{rgb}{0,0.6,0}
\definecolor{codegray}{rgb}{0.5,0.5,0.5}
\definecolor{codepurple}{rgb}{0.58,0,0.82}
\definecolor{backcolour}{rgb}{0.95,0.95,0.92}

\lstdefinestyle{mystyle}{
    commentstyle=\color{codegreen},
    keywordstyle=\color{magenta},
    numberstyle=\tiny\color{codegray},
    stringstyle=\color{codepurple},
    basicstyle=\fontsize{8}{9}\selectfont\ttfamily,
    breakatwhitespace=false,         
    breaklines=true,                 
    captionpos=b,                    
    keepspaces=true,                 
    numbers=left,                    
    numbersep=5pt,                  
    showspaces=false,                
    showstringspaces=false,
    showtabs=false,                  
    tabsize=2
}

\newsavebox{\mybox}

\title{Free Performance Gain from Mixing Multiple Partially Labeled Samples in Multi-label Image Classification}

\author{%
  Chak Fong Chong, Jielong Guo, Xu Yang, Wei Ke, Yapeng Wang\thanks{Corresponding author.} \\
  Faculty of Applied Sciences, Macao Polytechnic University, Macao SAR, China\\
  \texttt{\{chakfong.chong, jielong.guo, xuyang, wke, yapengwang\}@mpu.edu.mo} \\
}

\begin{document}

\maketitle

\begin{abstract}
Multi-label image classification datasets are often partially labeled where many labels are missing, posing a significant challenge to training accurate deep classifiers.
However, the powerful Mixup sample-mixing data augmentation cannot be well utilized to address this challenge, as it cannot perform linear interpolation on the unknown labels to construct augmented samples.
In this paper, we propose LogicMix, a Mixup variant designed for such partially labeled datasets.
LogicMix mixes the sample labels by logical OR so that the unknown labels can be correctly mixed by utilizing OR's logical equivalences, including the domination and identity laws.
Unlike Mixup, which mixes exactly two samples, LogicMix can mix multiple ($\geq2$) partially labeled samples, constructing visually more confused augmented samples to regularize training.
LogicMix is more general and effective than other compared Mixup variants in the experiments on various partially labeled dataset scenarios.
Moreover, it is plug-and-play and only requires minimal computation, hence it can be easily inserted into existing frameworks to collaborate with other methods to improve model performance with a negligible impact on training time, as demonstrated through extensive experiments.
In particular, through the collaboration of LogicMix, RandAugment, Curriculum Labeling, and Category-wise Fine-Tuning, we attain state-of-the-art performance on MS-COCO, VG-200, and Pascal VOC 2007 benchmarking datasets.
The remarkable generality, effectiveness, collaboration, and simplicity suggest that LogicMix promises to be a popular and vital data augmentation method.
Code is available at \url{XXX://XXX.XXX/XXX}.
\end{abstract}

\section{Introduction}
\label{sec:introduction}

Multi-label classification (MLC) of images aims to recognize the presence of multiple categories in an image. It has been applied to various applications like medical image interpretation \cite{chong_correlations_2023}
.
In the last decade, significant progress has been made in training deep MLC models on conventional fully labeled datasets, where in the datasets, the presence or absence of each category in each image is exhaustively annotated by a positive or negative label.
Nevertheless, large-scale datasets are often \textit{partially labeled} due to the expensive annotation cost \cite{kuznetsova_open_2020, gupta_lvis_2019, pham_learning_2021}.
For each image, only the labels of some categories are annotated, and the labels of the rest of the categories are unknown.
The missing of some labels leads to decreased supervision signals, raising a significant challenge to train high-performance models for \textbf{MLC} with \textbf{p}artially \textbf{l}abeled datasets (MLC-PL).

Formally, consider a MLC-PL task with $C$ categories. We denote a partially labeled dataset as $\mathcal{D}=\{(\mathbf{x}^{(i)}, \mathbf{y}^{(i)})\}$.
Each sample $(\mathbf{x}, \mathbf{y}) \in \mathcal{D}$ consists of an image $\mathbf{x}$ and a label vector $\mathbf{y}\in \{0, 1, ?\}^{C}$.
The $c^{\text{th}}$ element $y_c \in \{0, 1, ?\}$ ($c=1, ..., C$) is the label of category $c$, which is assigned to be either $0$ (negative), $1$ (positive), or $?$ (unknown), representing the category $c$ in $\mathbf{x}$ is absence, presented, or unknown, respectively.
The goal is training a MLC model with the partially labeled dataset $\mathcal{D}$.

Previous works on MLC-PL have retained remarkable success.
Durand \etal \cite{durand_learning_2019} proposed a loss function to explicitly ignore the unknown labels in training.
Opposite in the ignoring, Assume Negative (AN) \cite{kim_large_2022, kundu_exploiting_2020, ben-baruch_multi-label_2022} treats all unknown labels as negative labels.
Several works improved AN to treat some unknown labels as soft/hard negative labels based on image and category correlations \cite{kundu_exploiting_2020}, label likelihood and label prior \cite{ben-baruch_multi-label_2022}, memorization effect in noisy labels \cite{kim_large_2022}, or the approximate probability distribution of the unknown labels \cite{zhang_simple_2021}.
BoostLU \cite{kim_bridging_2023} increases the logits for positive labels of models trained with AN.
Moreover, various learning frameworks have also been proposed to exploit correlations between images and categories \cite{huynh_interactive_2020, chen_structured_2022, chen_heterogeneous_2022}, blend instance/prototype-level representations \cite{pu_semantic-aware_2022, pu_dual-perspective_2024}, self-train based on saliency regularization \cite{wang_saliency_2023}, predict unknown labels \cite{durand_learning_2019}, or cast MLC-PL into a Positive-Unlabeled problem \cite{yuan_positive_2023}.
Besides, CFT \cite{chong_category-wise_2023, chong_category-wise_2024} uses known labels to fine-tune models trained with pseudo-labels and \cite{sun_dualcoop_2022, hu_dualcoop_2023, ding_exploring_2023} adapts vision-language models to perform MLC-PL.

Orthogonal to the aforementioned previous works, this paper aims to bring Mixup data augmentation \cite{zhang_mixup_2017} into MLC-PL. Mixup is a famous and powerful yet simple rule-based sample-mixing data augmentation method, which has remarkably succeeded in many classification tasks \cite{song_toward_2024, li_dividemix_2020, thulasidasan_mixup_2019}.
It constructs an augmented sample $(\mathbf{x}', \mathbf{y}')$ by taking the element-wise linear interpolation of two random samples $(\mathbf{x}^{(1)}, \mathbf{y}^{(1)}), (\mathbf{x}^{(2)}, \mathbf{y}^{(2)})$:
\begin{align}
    \mathbf{x}' &= \lambda\mathbf{x}^{(1)} + (1-\lambda)\mathbf{x}^{(2)} \label{eq:mixup_x}\\
    y'_c &= \lambda y_c^{(1)} + (1-\lambda) y_c^{(2)}, \text{ for }c=\{1, 2, \dots, C\} \label{eq:mixup_y}
\end{align}
where $\lambda \sim Beta(\alpha, \alpha) \in [0, 1]$ ($\alpha$ is a hyperparameter).

As can be seen in (\ref{eq:mixup_y}), the construction of $\mathbf{y}'$ works well for fully labeled samples ($\mathbf{y}\in \{0, 1\}^C$). However, in MLC-PL, samples are partially labeled ($\mathbf{y}\in \{0, 1, ?\}^C$).
As an unknown label $?$ is not a valid value for the arithmetic operation in the label mixing (\ref{eq:mixup_y}), Mixup does not support mixing partially labeled samples.
Therefore, Mixup cannot be directly utilized to address the challenge of MLC-PL.

Through the literature review, we have recognized two solutions that may effectively adapt Mixup to MLC-PL:
1) Mixup-PME \cite{dong_revisiting_2022} is the only existing solution in the literature, to our knowledge. It first replaces all unknown labels $?$ in $\mathbf{y}^{(1)}, \mathbf{y}^{(2)}$ with $0.5$ to construct $\mathbf{y}'$ then set $y'_c=?$ if $y^{(1)}_c=?$.
However, it was only tested on a medical image dataset \cite{wang_chestx-ray8_2017}. Its generality on other datasets is to be investigated;
2) Inspired by AN, by replacing all the unknown labels $?$ in $\mathbf{y}^{(1)}, \mathbf{y}^{(2)}$ with negative labels $0$, the two samples become fully labeled and can be mixed by Mixup. However, this solution produces many false negative labels, significantly harming the model training.
Noteworthy, our experiments show that these two solutions are less general and effective in different datasets and known label proportions scenarios (\autoref{tab:logicmix_comparison}).
In a nutshell, Mixup has not been effectively utilized to help tackle the challenge of MLC-PL.

Therefore, this paper proposes a Mixup variant that can be effectively utilized in MLC-PL, called \textbf{LogicMix}. We design LogicMix starting from Mixup.
First, the construction of $\mathbf{y}'$ (\ref{eq:mixup_y}) is altered so that $\mathbf{y}^{(1)}, \mathbf{y}^{(2)} \in \{0, 1, ?\}^C$ are mixed by logical OR $\lor$ \cite{wang_baseline_2019, gao_dynamic_2022}.
As the unknown labels $?$ in $\mathbf{y}^{(1)}, \mathbf{y}^{(2)}$ cannot be operated by $\lor$, we propose utilizing $\lor$'s logical equivalences to correctly mix the unknown labels, which ensures the constructed $\mathbf{y}'$ can accurately describe the augmented image $\mathbf{x}'$, \ie, noiseless.
Subsequently, unlike Mixup, which mixes exactly two samples, LogicMix can mix multiple ($\geq 2$) partially labeled samples. So the constructed $(\mathbf{x}', \mathbf{y}')$ becomes visually more confused, which may provide more regularization power to model training.
Lastly, as the constructed $(\mathbf{x}', \mathbf{y}')$ generally has fewer negative labels, we propose training models by alternating between augmented samples and original samples, where the latter aims to supply negative labels.

LogicMix can be easily inserted into existing learning frameworks to improve model performance to a higher level.
In particular, we insert LogicMix into an existing learning framework \cite{chong_category-wise_2024} and achieve state-of-the-art (SOTA) results on three benchmarking datasets (average mAP 84.2\%, 49.9\%, 93.7\% on MS-COCO \cite{lin_microsoft_2014}, VG-200 \cite{chen_structured_2022, krishna_visual_2017}, Pascal VOC 2007 \cite{everingham_pascal_2009}, surpass/equal the previous SOTAs by +0.5\% \cite{chong_category-wise_2024}, +0.7\% \cite{wang_saliency_2023}, +0.0(2)\% \cite{wang_saliency_2023}, respectively).
Moreover, LogicMix can collaborate well with different advanced methods (and their combinations), including RandAugment \cite{cubuk_randaugment_2020}, CL \cite{durand_learning_2019}, CFT \cite{chong_category-wise_2023, chong_category-wise_2024}, LL-R \cite{kim_large_2022}, SST \cite{chen_structured_2022}, and SARB \cite{pu_semantic-aware_2022}, as demonstrated through extensive experiments.
Apart from its excellent collaboration ability, LogicMix itself is also more effective and general than other Mixup variants in different datasets and known label proportions scenarios. As LogicMix mixes samples based on simple rules, it is easy-to-implement and only causes a negligible impact on the training time. Hence, we describe it as a "free performance gain".

\section{LogicMix}

In this section, we section-by-section describe how we extend Mixup to our proposed LogicMix, including mixing partially labeled samples (Section \ref{sec:mix_partial_labels_logical_or}), enabling mixing multiple partially labeled samples (Section \ref{sec:logicmix_combine_multiple}), and the alternative training between augmented samples and original samples (Section \ref{sec:logicmix_alternating}). Finally, we give the definition of LogicMix in Section \ref{sec:logicmix_definition}.

\subsection{Mixing Partially Labeled Samples}
\label{sec:mix_partial_labels_logical_or}

As illustrated in Section \ref{sec:introduction}, Mixup does not support mixing partially labeled samples ($\mathbf{y} \in \{0, 1, ?\}^C$), because an unknown label $?$ is not a valid value for the arithmetic operation in the construction of $\mathbf{y}'$ (\ref{eq:mixup_y}).
To tackle this problem, we first follow \cite{wang_baseline_2019, gao_dynamic_2022} to alter Mixup's construction of $\mathbf{y}'$ (\ref{eq:mixup_y}) to mix $\mathbf{y}^{(1)}, \mathbf{y}^{(2)}$ with logical OR $\lor$:
\begin{equation}
    y'_c = y^{(1)}_c \lor y^{(2)}_c,  \text{ for }c=\{1, 2, \dots, C\} \label{eq:mlmixup_y_revisit}
\end{equation} .
(\ref{eq:mlmixup_y_revisit}) constructs hard $\mathbf{y}' \in \{0, 1\}$, unlike the origin (\ref{eq:mixup_y}) that constructs soft $\mathbf{y}' \in [0, 1]$. The hard $\mathbf{y}' \in \{0, 1\}$ is considered to be more feasible for MLC that uses a fixed threshold to interpret the model predictions to produce classification results \cite{gao_dynamic_2022}.

The above alternation does not yet enable Mixup to support mixing partially labeled samples, as an unknown label $?$ is also not a valid value for $\lor$.
Therefore, subsequently, we propose utilizing $\lor$'s logical equivalences to correctly mix unknown labels $?$ in (\ref{eq:mlmixup_y_revisit}). The logical equivalences including
\begin{equation}
    \text{\textbf{Domination law}: } q \lor 1 \equiv 1 \lor q \equiv 1 \text{ ; ~~~~~\textbf{Identity law}: } q \lor 0 \equiv 0 \lor q \equiv q \label{eq:logical_equivalenes}
\end{equation}
. The logical equivalences hold no matter what the unknown $q \in \{0, 1\}$ is.
Therefore, we can consider all $?$ in (\ref{eq:mlmixup_y_revisit}) to be unknowns and use the logical equivalences to compute $y'_c$.
For instance, if $y^{(1)}_c = 1$ and $y^{(2)}_c = ?$, from the domination law, $y'_c = 1 \lor ? = 1 \lor q=1$; If $y^{(1)}_c = ?$ and $y^{(2)}_c = 0$, from the identity law, $y'_c= ? \lor 0 = q \lor 0 = q = ? $.
In the case of $y^{(1)}_c = y^{(2)}_c = ?$, since the value of OR of two unknowns $q, r$ ($q\lor r$) is unknown, $y'_c=? \lor ? = q \lor r = ?$.

By combining the above-proposed method to tackle unknown labels into (\ref{eq:mlmixup_y_revisit}), we enable Mixup to support mixing two partially labeled samples. Formally, given $\mathbf{y}^{(1)}, \mathbf{y}^{(2)} \in \{0, 1, ?\}^C$, $\mathbf{y}'_c\in \{0, 1, ?\}^C$ is constructed by:
\begin{equation}
    y'_c = \begin{cases}
        1 &, \text{if \phantom{else~~} } y^{(1)}_c = 1 \text{ or } y^{(2)}_c = 1\\
        0 &, \text{else if~~ } y^{(1)}_c = y^{(2)}_c = 0 \\
        ? &, \text{else }
    \end{cases}
    ~~~~~, c=\{1, \dots, C\}\text{.} \label{eq:mix_partial_label}
\end{equation}.
Although the constructed $\mathbf{y}'$ contains unknown labels, its positive and negative labels can correctly reflect the presence and absence of the corresponding categories in $\mathbf{x}'$, \ie, noiseless.

\subsection{Mixing Multiple ($\geq 2$) Samples}
\label{sec:logicmix_combine_multiple}
The previous section have enabled Mixup to mix \textit{two} partially labeled samples with logical OR.
In this section, we extend it to support mixing \textit{multiple} ($\geq2$) partially labeled samples.
Mixing more samples result in visually more confused $\mathbf{x}'$, which may provide stronger regularization to training.

Let $(\mathbf{x}^{(1)}, \mathbf{y}^{(1)}), \dots, (\mathbf{x}^{(K)}, \mathbf{y}^{(K)})$ be the $K\in[2, 3, \dots, \infty)$ partially labeled samples to be mixed. We construct $(\mathbf{x}', \mathbf{y}')$ as described in follows.

\paragraph{Construction of $\mathbf{x}'$}

We extend Mixup's construction of $\mathbf{x}'$ (\ref{eq:mixup_x}) to support mixing $\mathbf{x}^{(1)}, \dots, \mathbf{x}^{(K)}$:
\begin{equation}
    \mathbf{x}' = \frac{\lambda^{(1)}\mathbf{x}^{(1)}+\cdots+\lambda^{(K)}\mathbf{x}^{(K)}}{\lambda^{(1)}+\cdots+\lambda^{(K)}} = \frac{1}{\sum_{k=1}^{K}{\lambda^{(k)}}}\sum_{k=1}^{K}\lambda^{(k)}\mathbf{x}^{(k)} \label{eq:logicmix_x}
\end{equation}
where $\lambda^{(k)}$ is the weight of $\mathbf{x}^{(k)}$.
If we follow the original Mixup to randomize $\lambda^{(k)}$, some $\mathbf{x}^{(k)}$ may contribute to small to $\mathbf{x}'$. Therefore, we set all $\lambda^{(k)}$ equal, ensuring all images contribute significantly.

\paragraph{Construction of $\mathbf{y}'$}

We first define the construction in the case that the samples are fully labeled ($\mathbf{y}^{(k)}\in \{0, 1\}^C$).
Considering that $\mathbf{x}'$ contains all visual features of $\mathbf{x}^{(1)}, \dots, \mathbf{x}^{(K)}$, meaning categories presented in any of $\mathbf{x}^{(1)}, \dots, \mathbf{x}^{(K)}$ are presented in $\mathbf{x}'$. To ensure the constructed $\mathbf{y}'$ can correctly describe the presence or absence of the categories in $\mathbf{x}'$, we construct $\mathbf{y}'$ by extending (\ref{eq:mlmixup_y_revisit}):
\begin{equation}
    y'_c = y^{(1)}_c \lor \cdots \lor y^{(K)}_c,  \text{ for }c=\{1, 2, \dots, C\} \label{eq:mlmixup_y_multiple}
\end{equation}.
Then, we use our proposed method in Section \ref{sec:mix_partial_labels_logical_or} to enable the construction to support mixing partially labeled samples ($\mathbf{y}^{(k)}\in \{0, 1, ?\}^C$). If any one of $y^{(1)}_c, \dots, y^{(K)}_c$ in (\ref{eq:mlmixup_y_multiple}) is $1$, by applying the domination law $K-1$ times to (\ref{eq:mlmixup_y_multiple})-right, it is easy to compute $y'_c=1$;
Else, if all $y^{(1)}_c, \dots, y^{(K)}_c$ are $0$, it is easy to compute $y'_c=0$;
In the other cases where no $y^{(1)}_c, \dots, y^{(K)}_c$ is $1$ and at least one of $y^{(1)}_c, \dots, y^{(K)}_c$ is $?$, even if we apply the identity law multiple times to (\ref{eq:mlmixup_y_multiple})-right to eliminate all $0$, at least one $?$ remains, meaning $y'_c=?$. To sum up, $\mathbf{y}'$ is constructed by:
\begin{equation}
    y'_c = \begin{cases}
        1 &, \text{if } \exists k \in \{1, \dots, K\}y^{(k)}_c = 1 \\
        0 &, \text{else if } \forall k \in \{1, \dots, K\}y^{(k)}_c = 0 \\
        ? &, \text{else }
    \end{cases}
    ~~~~~, c=\{1, \dots, C\} \label{eq:logicmix_y} \text{.}
\end{equation}

In experiments, we set each augmented sample is constructed by mixing $K \sim \mathcal{U}\{K_{min}, K_{max}\}$ samples, where $K_{min}, K_{max}$ are hyperparameters.

\subsection{Alternating Between Augmented Samples and Original Samples}
\label{sec:logicmix_alternating}

In the above-proposed mixing of multiple partially labeled samples (\ref{eq:logicmix_y}), the positive/negative labels of $\mathbf{y}'$ are the \textit{union}/\textit{intersection} of the positive/negative labels of $\mathbf{y}^{(1)}, ..., \mathbf{y}^{(K)}$, respectively. This results in $\mathbf{y}'$ generally having more positive labels, but fewer negative samples.
For instance, in the MS-COCO dataset with a known label proportion 30\%, each sample has mean 0.88/23.1 positive/negative labels. If we apply (\ref{eq:logicmix_x}) and (\ref{eq:logicmix_y}) to mix multiple samples ($K_{min}=2, K_{max}=3$) to construct augmented samples, each augmented sample has mean 2.11/4.32 positive/negative labels, which has 2.41x more positive labels but 5.34x fewer negative labels.
The more positive labels benefit model training, as positive labels are rarely seen under the dominance of negative labels in most MLC datasets. Yet, the less negative labels are worth being addressed.

To alleviate this problem, we propose training the model by alternating between augmented samples and original samples. Specifically, a sample being fed into the model has a $s\times 100\%$ chance of being an augmented sample and has a $(1-s)\times 100\%$ chance of being an original sample (where $s$ is a hyperparameter).
The use of original samples is to supply negative labels and moreover, ensures the model has seen original images similar to the images in the evaluation, instead of only the augmented images.

\subsection{LogicMix Definition}
\label{sec:logicmix_definition}
\begin{figure}
    \centering
    \begin{tikzpicture}
        \node (img) at (0, 0) {\includegraphics[width=\linewidth]{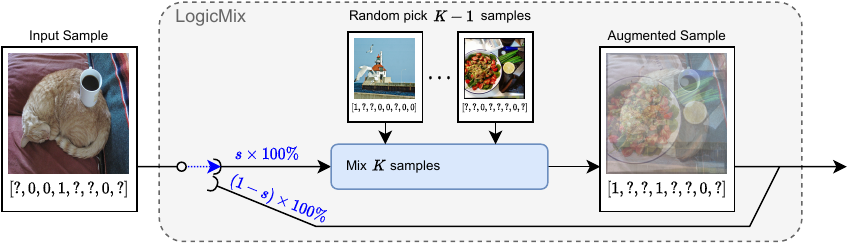}};
        \node at (29pt, -20pt) {\scriptsize Eq. (\ref{eq:logicmix_x}, \ref{eq:logicmix_y})};
    \end{tikzpicture}
    \vspace{-16pt}
    \caption{The overview of LogicMix.}
    \label{fig:logic_mix}
\end{figure}

The proposed LogicMix is the integration of mixing multiple partially labeled samples (Section \ref{sec:logicmix_combine_multiple}) and the alternating between augmented samples and original samples (Section \ref{sec:logicmix_alternating}). We present LogicMix in the form of a data augmentation pipeline as follows and in \autoref{fig:logic_mix}.

\paragraph{LogicMix} Before each sample $(\mathbf{x}, \mathbf{y}) \in \mathcal{D}$ is used for training, it goes through the LogicMix data augmentation pipeline. Inside LogicMix, the input sample has a probability of $s\times 100\%$ being augmented. In the case of augmentation, the pipeline first draws a random number $K \sim \mathcal{U}\{K_{min}, K_{max}\}$ that determines the number of samples to be mixed. Subsequently, the pipeline randomly picks $K-1$ samples from $\mathcal{D}$. Finally, the input sample and the $K-1$ samples are mixed by (\ref{eq:logicmix_x}) and (\ref{eq:logicmix_y}) to construct the augmented sample, the output of LogicMix. In the other case (no augmentation), the output of LogicMix is identical to the input sample.

\begin{table}[]
    \centering
    \footnotesize
    \caption{Comparison of LogicMix to Mixup variants for MLC.}
    \vspace{6pt}
    \label{tab:logicmix_feature_comparison}
    \begin{tabular}{lcccccc}
        \hline
         \textbf{Method} & \makecell{Support Partially \\ Labeled Samples} & \makecell{Construct Hard\\ Augmented Labels} & \makecell{Mix $\geq2$ \\ Samples} & \makecell{Alternative Training w/ \\Original Samples} \\
         \hline
         Mixup \cite{zhang_mixup_2017} & \ding{53} & \ding{53} & \ding{53} & \ding{53} \\
         Wang \etal \cite{wang_baseline_2019} & \ding{53} & \ding{51} & \ding{53} & \ding{51} \\
         Multi-label Mixup \cite{gao_dynamic_2022} & \ding{53} & \ding{51} & \ding{53} & \ding{53} \\
         Mixup-PME \cite{dong_revisiting_2022} & \ding{51} & \ding{53} & \ding{53} & \ding{53} \\
         \textbf{LogicMix (Ours)} & \ding{51}& \ding{51} & \ding{51} & \ding{51} \\
         \hline
    \end{tabular}
    
\end{table}

\paragraph{Comparison to Mixup Variants for MLC} We compare the designs of LogicMix to other Mixup variants suitable for MLC, as summarized in \autoref{tab:logicmix_feature_comparison}. LogicMix first supports mixing partially labeled samples to construct augmented samples with hard labels, where the hard labels are considered to be more feasible for MLC \cite{gao_dynamic_2022}. LogicMix also first supports mixing more than two partially labeled samples, to the best of our knowledge. This comparison highlights the novelties of our LogicMix. Besides, we conduct experimental comparisons in Section \ref{sec:logicmix_comparison_exp}.

\paragraph{Impact to Training Time} LogicMix introduces some disk I/O to load samples and minimal CPU computation to mix the samples by arithmetic mean (\ref{eq:logicmix_x}) and logical operations (\ref{eq:logicmix_y}) in the data preprocessing stage. The impact on the training time can be minimized to be negligible by introducing more data loader workers, as reported in Appendix \ref{sec:computation_time}.

\section{Learning Framework with LogicMix}

To demonstrate that LogicMix can effectively collaborate with recent advanced methods, we insert LogicMix into the state-of-the-art learning framework on the MS-COCO dataset that was obscurely proposed in \cite{chong_category-wise_2024}.
The framework mainly consists of three recent advanced methods. (1) \textbf{RandAugment} \cite{cubuk_randaugment_2020} randomly applies single image transformations to each sample; (2) \textbf{Curriculum Learning (CL)} \cite{durand_learning_2019} gradually generates pseudo-labels for easy unknown labels; And (3) \textbf{Category-wise Fine-Tuning (CFT)} \cite{chong_category-wise_2023, chong_category-wise_2024} one-by-one fine-tunes the trained model's classification heads using only known labels.
The following sections describe the framework with LogicMix (a visual illustration is available in Appendix \ref{sec:framework_visual}). The other implementation details (model architecture, loss functions, etc.) can be found in Section \ref{sec:experimental_setups}.

\subsection{Data Augmentation with RandAugment and LogicMix}

Each sample is sequentially augmented by RandAugment and LogicMix.
RandAugment \cite{cubuk_randaugment_2020} is an automatic data augmentation method that automatically selects transformations to apply to the input sample. RandAugment has two hyperparameters, $N$ and $M$. $N$ is the number of transformations applied to each sample. $M$ is the magnitude of all the transformations. For each sample, RandAugment randomly picks $N$ transformations from the following 14 transformations:
Identity, AutoContrast, Equalize, Rotate, Solarize, Color, Posterize, Contrast, Brightness, Sharpness, Shear-X, Shear-Y, Translate-X, and Translate-Y.
Each transformation has the same probability of being picked. The picked $N$ transformations with magnitude $M$ are sequentially applied to the input sample. 
This collaboration of LogicMix with RandAugment is to demonstrate that LogicMix can collaborate with different data augmentation methods (the methods included in RandAugment) effectively. 

\subsection{Curriculum Labeling}
CL \cite{durand_learning_2019} is a curriculum learning-based method \cite{bengio_curriculum_2009} that gradually generates pseudo-labels for easy unknown labels during training. The pseudo-labels are added to the training data and participate with known labels in the rest of the training.

At the end of each epoch, the model predicts the logit (before Sigmoid) $p^{(i)}_c \in \mathbb{R}$ for each unknown label $y^{(i)}_c=?$ in the dataset. If $p^{(i)}_c$ is greater than the positive threshold for the positive label $\theta^+$ or less than the threshold for the negative label $\theta^-$, the model is confident in its prediction (\ie, easy unknown label). In this case, we generate a pseudo-label $\bar{y}^{(i)}_c$ for the unknown label $y^{(i)}_c$:
\begin{equation}
    \bar{y}^{(i)}_c = \begin{cases}
        1 &, p^{(i)}_c > \theta^+ \\
        0 &, p^{(i)}_c < \theta^-
    \end{cases}
\end{equation} .
The pseudo-label $\bar{y}^{(i)}_c$ replaces the unknown label $y^{(i)}_c$ only in the next epoch. At the end of the next epoch, $\bar{y}^{(i)}_c$ is discarded, and a new pseudo-label will be generated.

Unlike the original paper \cite{durand_learning_2019} and \cite{chong_category-wise_2024}, which use a single threshold $\theta$ to determine the generations of both positive and negative pseudo-labels, we decouple $\theta$ into two thresholds $\theta^+, \theta^-$, providing more freedom to control the difficulties of positive and negative pseudo-label generations.

\subsection{Category-wise Fine-Tuning}
CL may generate wrong pseudo-labels that harm the training, as neural network models inevitably produce wrong predictions. To this end, CFT \cite{chong_category-wise_2023, chong_category-wise_2024} is employed to reduce the impact of the wrong pseudo-labels. CFT is applied after the model is well-trained. It one-by-one fine-tunes the model's classification heads using only known labels, which can calibrate the model's output to reduce the impact of wrong pseudo-labels during training. The details can be found in \cite{chong_category-wise_2023, chong_category-wise_2024}.

\section{Experiments}
We first describe the experimental setups (Section \ref{sec:experimental_setups}), then report the results of the learning framework with LogicMix (Section \ref{sec:result_framework}), followed by the results of LogicMix (Section \ref{sec:result_logicmix}).

\subsection{Experimental Setups}
\label{sec:experimental_setups}
\paragraph{Datasets and Performance Metric} Following the previous works \cite{chen_structured_2022, chen_heterogeneous_2022, pu_semantic-aware_2022, pu_dual-perspective_2024}, we evaluate our methods on three benchmarking datasets, including \textbf{MS-COCO} \cite{lin_microsoft_2014}, \textbf{VG-200} \cite{chen_structured_2022}, and \textbf{Pascal VOC 2007} \cite{everingham_pascal_2009}. More details can be found in Appendix \ref{sec:experimental_setups_detail}.
Since the three datasets are fully labeled, we follow previous works to simulate partially labeled datasets with known label proportions 10\%, 20\%, ..., 90\% by randomly dropping 90\%, 80\%, ..., 10\% of labels. We report the model performance in mean average precision (mAP).

\paragraph{Implementation Details}
\label{sec:implementation_details}

For fair comparisons to previous work, ResNet-101 \cite{he_deep_2016} pre-trained on ImageNet-1K \cite{deng_imagenet_2009} is initialized to be the model backbone. A multi-layer perception is appended on the backbone to output the classification predictions.
The model is trained for 60 epochs.
Partial asymmetric loss \cite{ben-baruch_multi-label_2022} ($\gamma^+=4, \gamma^-=0,$ and $m=0.05$) is used to compute the loss.
The batch size is set to 32, with a gradient accumulation of 4 batches (\ie, global batch size 128).
Adam optimizer \cite{kingma_adam_2014} is used to update parameters.
The initial learning rate is set to be $2\times 10^{-4}$, decayed by 1cycle policy \cite{smith_super-convergence_2019}.
True weight decay with $1\times 10 ^{-4}$ is used, and the model's exponential moving average is maintained.
Images are resized to $448^2$ and rescaled to range $[0, 1]$.
For RandAugment, we set $N=2$ and $M=9$.
For CFT, we choose to optimize each classification head's asymmetric loss \cite{ridnik_asymmetric_2021} with the same configurations as in the original paper \cite{chong_category-wise_2024}.
We set different hyperparameters of LogicMix ($p, K_{min}, K_{max}$) and CL ($\theta^+, \theta^-$) for each dataset.
For MS-COCO, we set $s=0.5, K_{min}=2, K_{max}=3, \theta^+=2, \theta^-=-2$.
For VG-200, we set $s=0.5, K_{min}=2, K_{max}=3, \theta^+=2, \theta^-=-1$.
For Pascal VOC 2007, we set $s=0.5, K_{min}=3, K_{max}=4, \theta^+=2.5, \theta^-=-1$.

\subsection{Results of the Learning Framework with LogicMix}
\label{sec:result_framework}

\subsubsection{Comparison to State-of-the-Art (SOTA)}

\autoref{tab:sota} compares of our framework with LogicMix to SOTAs introduced in the following.
\textbf{SSGRL} \cite{chen_learning_2019}, \textbf{GCN-ML} \cite{chen_multi-label_2019}, \textbf{KGGR} \cite{chen_knowledge-guided_2020}, and \textbf{P-GCN} \cite{chen_learning_2021} employ graph neural networks to exploit relationships between categories for fully labeled datasets. They were adapted \cite{chen_heterogeneous_2022} for MLC-PL by replacing the original losses with the partial-BCE loss \cite{durand_learning_2019}.
\textbf{PLCL} \cite{durand_learning_2019} gradually generates pseudo-labels for easy unknown labels.
\textbf{SPLC} \cite{zhang_simple_2021} treats unknown labels as negative labels and gradually corrects false negative labels based on an approximate distribution of unknown labels.
\textbf{SST} \cite{chen_structured_2022} and \textbf{HST} \cite{chen_heterogeneous_2022} frameworks exploit within-image and cross-image semantic correlations to generate pseudo-labels.
\textbf{SARB} \cite{pu_semantic-aware_2022} and \textbf{PSRB} \cite{pu_dual-perspective_2024} frameworks blend representations across samples to complement unknown labels.
\textbf{CSL} \cite{ben-baruch_multi-label_2022} treats some unknown labels as negative labels based on label likelihood and label prior.
\textbf{SR} \cite{wang_saliency_2023} models the saliency of the class-specific maps and strengthens the saliency of object regions corresponding to the presented labels.
\textbf{PU-MLC} \cite{yuan_positive_2023} casts the MLC-PL problem as a Positive-Unlabeled problem.
\textbf{CFT} \cite{chong_category-wise_2023, chong_category-wise_2024} uses known labels to fine-tune the well-trained model's classification heads. 

\begin{table}[]
    \centering
    \footnotesize
    \scriptsize
    \caption{Results of the proposed framework with LogicMix and comparison to state-of-the-art, in mAP\%. (Best is \textbf{bold} and second-best is \underline{underlined})}
    \vspace{6pt}
    \label{tab:sota}
    \begin{tabular}{clcccccccccc}
        \hline
        \multirow{2}{*}{Dataset} & \multirow{2}{*}{Method} & \multicolumn{9}{c}{Known Label Proportion} & \multirow{2}{*}{Average} \\
         & & 10\% & 20\% & 30\% & 40\% & 50\% & 60\% & 70\% & 80\% & 90\% & \\
         \hline
         \multirow{15}{*}{MS-COCO}
         & SSGRL \cite{chen_learning_2019} & 62.5 &70.5 &73.2 &74.5 &76.3 &76.5 &77.1 &77.9 &78.4 & 74.1 \\
         & GCN-ML \cite{chen_multi-label_2019} & 63.8 &70.9 &72.8 &74.0 &76.7 &77.1 &77.3 &78.3 &78.6 &74.4 \\
         & KGGR \cite{chen_knowledge-guided_2020}& 66.6 & 71.4 & 73.8 & 76.7 & 77.5 & 77.9 & 78.4 & 78.7 & 79.1 & 75.6 \\
         & P-GCN \cite{chen_learning_2021}& 67.5 & 71.6 & 73.8 & 75.5 & 77.4 & 78.4 & 79.5 & 80.7 & 81.5 & 76.2 \\
         & PLCL \cite{durand_learning_2019} &68.8&72.8&74.6&75.9&76.6&77.3&77.7&78.0&78.1&75.5 \\
         &SPLC \cite{zhang_simple_2021}&64.1&72.0&75.3&76.7&77.1&77.5&77.3&76.9&76.9&74.9\\
        &SST \cite{chen_structured_2022}&68.1&73.5&75.9&77.3&78.1&78.9&79.2&79.6&79.9&76.7\\
        &HST \cite{chen_heterogeneous_2022}&70.6&75.8&77.3&78.3&79.0&79.4&79.9&80.2&80.4&77.9\\
        &SARB \cite{pu_semantic-aware_2022}&71.2&75.0&77.1&78.3&78.9&79.6&79.8&80.5&80.5&77.9\\
        &PSRB \cite{pu_dual-perspective_2024}&72.5&76.0&77.6&78.7&79.6&79.8&80.0&80.5&80.8&78.4\\
        &CSL \cite{ben-baruch_multi-label_2022}&67.0&72.4&74.9&76.6&77.7&78.6&79.3&79.8&80.3&76.3\\
        & PU-MLC \cite{yuan_positive_2023} & 75.7 & 78.6 & 80.2 & 81.3 & 82.0 & 82.6 & 83.0 & 83.5 & 83.8 & 81.2 \\
        &SR \cite{wang_saliency_2023} &77.2&79.2&80.3&80.9&81.8&82.1&82.1&82.6&82.7&81.0\\
        &CL+CFT \cite{chong_category-wise_2024}&\underline{78.4}&\underline{81.9}&\underline{83.2}&\underline{84.2}&\underline{84.6}&\underline{84.9}&\underline{85.1}&\underline{85.3}&\underline{85.6}&\underline{83.7}\\
        &\textbf{Ours} & \textbf{79.0} & \textbf{82.2} & \textbf{83.5} & \textbf{84.3} & \textbf{85.0} & \textbf{85.4} & \textbf{85.8} & \textbf{86.1} & \textbf{86.2} & \textbf{84.2} \\
        \hline
        \multirow{15}{*}{VG-200}
        &SSGRL \cite{chen_learning_2019}&34.6&37.3&39.2&40.1&40.4&41.0&41.3&41.6&42.1&39.7\\
        &GCN-ML \cite{chen_multi-label_2019}&32.0&37.8&38.8&39.1&39.6&40.0&41.9&42.3&42.5&39.3\\
        & KGGR \cite{chen_knowledge-guided_2020} & 36.0 & 40.0 & 41.2 & 41.5 & 42.0 & 42.5 & 43.3 & 43.6 & 43.8 & 41.5\\
        & P-GCN \cite{chen_learning_2021} & - & - & - & - & - & - & - & - & - & - \\
        &PLCL \cite{durand_learning_2019} &40.5&42.9&43.9&44.5&44.9&45.0&45.1&45.2&45.3&44.1\\
        &SPLC \cite{zhang_simple_2021}&39.5&42.8&45.0&45.9&46.4&46.6&46.6&46.5&46.3&45.1\\
        &SST \cite{chen_structured_2022}&38.8&39.4&41.1&41.8&42.7&42.9&43.0&43.2&43.5&41.8\\
        &HST \cite{chen_heterogeneous_2022}&40.6&41.6&43.3&44.6&45.2&45.8&46.8&47.2&47.8&44.8\\
        &SARB \cite{pu_semantic-aware_2022}&40.6&43.5&44.5&45.3&46.0&47.1&47.2&47.8&48.1&45.6\\
        &PSRB \cite{pu_dual-perspective_2024}&41.4&44.0&44.8&45.5&46.6&47.5&47.8&48.0&48.2&46.0\\
        &CSL \cite{ben-baruch_multi-label_2022}&40.7&43.7&45.2&46.2&46.8&47.3&47.8&48.2&48.5&46.0\\
        & PU-MLC \cite{yuan_positive_2023} &-&-&-&-&-&-&-&-&-&-\\
        &SR \cite{wang_saliency_2023}&\textbf{46.7}&\textbf{48.2}&\underline{49.0}&\underline{49.5}&\underline{49.8}&\underline{49.9}&\underline{50.0}&\underline{50.1}&\underline{50.2}&\underline{49.2}\\
        &CL+CFT \cite{chong_category-wise_2024}&-&-&-&-&-&-&-&-&-&-\\
        &\textbf{Ours} & \underline{45.9} & \underline{48.1} & \textbf{49.2} & \textbf{50.0} & \textbf{50.5} & \textbf{50.9} & \textbf{51.3} & \textbf{51.6} & \textbf{51.8} & \textbf{49.9} \\
        \hline
        \multirow{15}{*}{\makecell{Pascal \\VOC 2007}}
        &SSGRL \cite{chen_learning_2019}&77.7&87.6&89.9&90.7&91.4&91.8&92.0&92.2&92.2&89.5\\
        &GCN-ML \cite{chen_multi-label_2019}&74.5&87.4&89.7&90.7&91.0&91.3&91.5&91.8&92.0&88.9\\
        & KGGR \cite{chen_knowledge-guided_2020} & 81.3 & 88.1 & 89.9 & 90.4 & 91.2 & 91.3 & 91.5 & 91.6 & 91.8 & 89.7 \\
        & P-GCN \cite{chen_learning_2021} & 82.5 & 85.4 & 88.2 & 89.8 & 90.0 & 90.9 & 91.6 & 92.5 & 93.1 & 89.3 \\
        &PLCL \cite{durand_learning_2019}&87.0&90.8&92.2&92.9&93.4&93.5&93.7&93.9&94.0&92.4\\
        &SPLC \cite{zhang_simple_2021}&79.4&86.2&89.9&91.8&92.6&92.9&93.3&93.6&93.7&90.4\\
        &SST \cite{chen_structured_2022}&81.5&89.0&90.3&91.0&91.6&92.0&92.5&92.6&92.7&90.4\\
        &HST \cite{chen_heterogeneous_2022}&84.3&89.1&90.5&91.6&92.1&92.4&92.5&92.8&92.8&90.9\\
        &SARB \cite{pu_semantic-aware_2022}&83.5&88.6&90.7&91.4&91.9&92.2&92.6&92.8&92.9&90.7\\
        &PSRB \cite{pu_dual-perspective_2024}&85.7&89.8&91.8&92.0&92.3&92.7&92.9&93.1&93.2&91.5\\
        &CSL \cite{ben-baruch_multi-label_2022}&85.4&89.3&91.3&92.3&92.6&93.2&93.6&93.8&94.0&91.7\\
        &PU-MLC \cite{yuan_positive_2023} & 88.0 & 90.7 & 91.9 & 92.0 & 92.4 & 92.7 & 93.0 & 93.4 & 93.5 & 92.0 \\
        &SR \cite{wang_saliency_2023}&\textbf{91.3}&\textbf{93.1}&\textbf{93.6}&\textbf{94.0}&\underline{94.1}&\underline{94.3}&\underline{94.3}&\underline{94.3}&\underline{94.3}&\textbf{93.7}\\
        &CL+CFT \cite{chong_category-wise_2024}&-&-&-&-&-&-&-&-&-&-\\
        &\textbf{Ours} & 90.7 & 92.5 & 93.4 & 93.8 & \textbf{94.2} & \textbf{94.4} & \textbf{94.8} & \textbf{94.8} & \textbf{94.9} & \textbf{93.7} \\
        \hline
    \end{tabular}
\end{table}

\paragraph{Performance on MS-COCO}
MS-COCO is one of the most widely used datasets for performance comparison, hence the results are more significant and representative. Our framework achieves average mAP 84.2\%, surpassing all compared methods. Our framework also dominates others on all tested known label proportions 10\%, 20\%, ..., 90\%. CL+CFT is the previous state-of-the-art, which is also the origin of our framework with LogicMix. We successfully improves the average mAP from 83.7\% to new SOTA 84.2\% (+0.5\%). A more detailed analysis is provided in the next section.

\paragraph{Performance on VG-200}
Our framework attains average mAP 49.9\%, outperforming all compared methods, including the previous SOTA SR (+0.7\%).
In deeper comparison to SR, we achieve higher mAPs on known label proportions 30\%, 40\%, ..., 90\%.
As the known label proportion grows from 10\% to 90\%, SR improves mAP from 46.7\% to 50.2\% only (+3.5\%), while our framework significantly improves mAP from 45.9\% to 51.8\% (+5.9\%).
This result might imply that our framework can better transform labels to performance than SR.

\paragraph{Performance on Pascal VOC 2007} 
As the SOTA SR has already achieved average mAP 93.7\%, it is more difficult to improve further.
Our framework achieves average mAP 93.72\%, which may be better than the SOTA SR 93.7?\%, at least equal.
In deeper comparison to SR, our framework outperforms SR on known label proportions $\geq$ 50\%. Especially, SR achieves the same mAPs 94.3\% on proportions 60\% to 90\%, meaning it may inefficiently utilize the additional 30\% of labels. In contrast, our framework improves mAP from 94.4\% to 94.9\%, which again demonstrates our framework might be better at transforming labels to performance.

\subsubsection{Framework Analysis}

\autoref{tab:framework_ablation} shows the ablation studies of our framework on MS-COCO.
The single uses of LogicMix, RandAugment, and CL improve average mAP by +1.24\%, +1.32\%, and +0.48\%, respectively.
CFT improves the single use of CL by +0.14\% (CFT must collaborate with CL).
The combined uses of any four methods are always better than their subsets in average mAP.

Here, we highlight the contribution of LogicMix to the framework.
LogicMix consistently improves the mAPs of the combined uses of the other three components (or their subsets) by +0.27\% $\sim$ +1.01\%,  +0.68\% $\sim$ +1.46\%, +0.94\% $\sim$ +1.84\%, on tested proportions 10\%, 50\%, 90\%, respectively. These results suggest LogicMix can collaborate well with RandAugment, CL, CFT, and their combinations.

\paragraph{Collaboration with Other Methods and the Results with Other Model Architectures} 
Besides the three methods used in the framework, we further demonstrate LogicMix's collaboration with SST \cite{chen_structured_2022}, SARB \cite{pu_semantic-aware_2022}, and LL-R \cite{kim_large_2022}, as detailed in Appendix \ref{sec:cooperation_others}. We also report the results with different model architectures, including TResNet-M \cite{ridnik_tresnet_2021} and SwinTransformerV2-Base \cite{liu_swin_2022} in Appendix \ref{sec:framework_architecture}.

\begin{table}[]
    \centering
    \scriptsize
    \caption{Ablation study of the framework in MS-COCO, in mAP\%. "Avg." is the average mAP of proportions 10\%, 50\%, and 90\%.}
    \vspace{6pt}
    \label{tab:framework_ablation}
    \begin{tabular}{cccccccc}
         \hline
         LogicMix & RandAugment & CL & CFT & 10\% & 50\% & 90\% & Avg. \\
         \hline
                    &            &            &            & 75.7 & 82.0 & 83.4 & 80.4 \\
         \checkmark &            &            &            & \textbf{76.3} & \textbf{83.5} & \textbf{85.1} & \textbf{81.6} \\
         \hline
                    & \checkmark &            &            & 76.79 & 83.3 & 84.9 & 81.7 \\
         \checkmark & \checkmark &            &            & \textbf{77.4} & \textbf{84.4} & \textbf{86.3} & \textbf{82.7} \\
         \hline
                    &            & \checkmark &            & 76.6 & 82.6 & 83.3 & 80.8 \\
         \checkmark &            & \checkmark &            & \textbf{77.2} & \textbf{83.7} & \textbf{85.0} & \textbf{82.0} \\
         \hline
                    &            & \checkmark & \checkmark & 76.8 & 82.8 & 83.4 & 81.0 \\
         \checkmark &            & \checkmark & \checkmark & \textbf{77.8} & \textbf{84.0} & \textbf{85.3} & \textbf{82.4} \\
         \hline
                    & \checkmark & \checkmark &            & 78.2 & 84.1 & 85.1 & 82.5 \\
         \checkmark & \checkmark & \checkmark &            & \textbf{78.5} & \textbf{84.8} & \textbf{86.1} & \textbf{83.1} \\
         \hline
                    & \checkmark & \checkmark & \checkmark & 78.4 & 84.2 & 85.3 & 82.6 \\
         \checkmark & \checkmark & \checkmark & \checkmark & \textbf{79.0} & \textbf{85.0} & \textbf{86.2} & \textbf{83.4} \\
         \hline       
    \end{tabular}
\end{table}

\subsection{Results of LogicMix}
\label{sec:result_logicmix}

\paragraph{Comparison to Mixup Variants}
\label{sec:logicmix_comparison_exp}

We implement some relevant Mixup variants suitable for MLC and compare them to LogicMix.
The variants include the original \textbf{Mixup} \cite{zhang_mixup_2017}, \textbf{the variant proposed by Wang \etal} \cite{wang_baseline_2019}, and \textbf{Multi-label Mixup} \cite{gao_dynamic_2022}.
Since these variants do not support mixing partially labeled samples, we use AN to replace each unknown label with a negative label, which turns the samples fully labeled.
We also compare to \textbf{Mixup-PME} \cite{dong_revisiting_2022}, which treats unknown labels as $0.5$.
Besides, we compare to models trained with \textbf{no augmentation} as the baseline.
The detailed introduction and the implementation details can be found in Appendix \ref{sec:implementation_details_other_mixup}.

\autoref{tab:logicmix_comparison} reports the results.
On MS-COCO and VG-200, LogicMix surpasses all the compared variants on all tested known label proportions.
On Pascal VOC 2007, LogicMix performs comparable to Mixup-PME, and overall better than the other three variants.
It is important to note that our LogicMix is the only one that always surpasses no augmentation on all tested scenarios. Mixup, Wang \etal, and ML-Mixup (AN) often decrease the mAPs on known label proportions of 10\% and 50\% on all three datasets, probably because AN generates many false negative labels harming the model training. Mixup-PME decreases the mAPs on proportions 10\% and 50\% on MS-COCO. The results suggest LogicMix is more general and effective than the compared Mixup variants.

\begin{table}[]
    \centering
    \scriptsize
    \caption{Comparison of LogicMix to other Mixup-like methods, in mAP\%. (Best is \textbf{bold} and second-best is \underline{underlined}. Results worse than no augmentation are marked in \color{red}{red})}
    \vspace{6pt}
    \label{tab:logicmix_comparison}
    \begin{tabular}{lccccccccc}
        \hline
        \multirow{2}{*}{Method} & \multicolumn{3}{c}{MS-COCO} & \multicolumn{3}{c}{VG-200} & \multicolumn{3}{c}{Pascal VOC 2007}\\
        \cline{2-10}
         & 10\% & 50\% & 90\% & 10\% & 50\% & 90\% & 10\% & 50\% & 90\% \\
         \hline
         No augmentation & \underline{75.7} & \underline{82.0} & 83.4 & 41.0 & 46.8 & 48.3 & 84.5 & 91.7 & 92.9 \\
         Mixup \cite{zhang_mixup_2017} (AN) & \color{red}{62.8} & \color{red}{77.2} & \color{red}{82.1} & \color{red}{37.0} & \color{red}{44.9} & 48.5 & \color{red}{77.6} & \color{red}{86.4} & \color{red}{92.2} \\
         Wang \etal \cite{wang_baseline_2019} (AN) & \color{red}{67.2} & \color{red}{79.5} & \underline{84.3} & \color{red}{39.3} & 47.1 & \underline{50.0} & \color{red}{76.5} & \color{red}{88.8} & \textbf{93.6}\\
         Multi-label Mixup \cite{gao_dynamic_2022} (AN) & \color{red}{68.1} & \color{red}{79.0} & 83.5 & \color{red}{39.8} & 47.0 & {49.6} & \color{red}{78.1} & \color{red}{87.5} & 92.9 \\
         Mixup-PME \cite{dong_revisiting_2022} & \color{red}{74.2} & \color{red}{81.6} & {83.6} & \underline{41.8} & \underline{47.7} & 49.3 & \textbf{86.4} & \underline{92.5} & {93.6} \\
         \textbf{LogicMix (Ours)} & \textbf{76.3} & \textbf{83.5} & \textbf{85.1} & \textbf{42.5} & \textbf{48.5} & \textbf{50.1} & \underline{85.5} & \textbf{92.5} & \underline{93.6} \\
         \hline
         
    \end{tabular}
    
\end{table}

\paragraph{Ablation Studies}
We validate the designs of LogicMix, including \textbf{m}ixing partially labeled samples using \textbf{l}ogical \textbf{e}quivalences (\textbf{MLE}), \textbf{m}ixing \textbf{m}ultiple \textbf{s}amples (\textbf{MMS}), and the alternate training between \textbf{a}ugmented samples and \textbf{o}riginal samples (\textbf{A}$\mathbf{\rightleftharpoons}$\textbf{O}).
\autoref{tab:logicmix_ablation} shows the results.
MLE consistently improves average mAPs by +1.1\%$\sim$+2.4\%, showing that mixing partially labeled samples by logical equivalences is better than treating unknown labels as negative labels.
A$\rightleftharpoons$O also consistently improves average mAPs by +2.0\%$\sim$+3.9\%.
MMS needs to cooperate with A$\rightleftharpoons$O to improve mAP.
Overall, the combination of all three designs attains the highest mAPs.

\paragraph{Hyperparameters Analysis} We vary LogicMix's hyperparameters $K_{min}, K_{max}, s$ to analyze their impacts. First we fix $s=0.5$ and vary $K_{min}, K_{max}$. The results in Figure \ref{fig:hyparam} (left) show that mixing multiple samples (\eg, $K_{min}=2,  K_{max}=3$) can outperform mixing exactly two samples ($K_{min}=K_{max}=2$), again showing the effectiveness of mixing multiple samples. Then we fix $K_{min}=2,  K_{max}=3$ and vary $s$. As shown in Figure \ref{fig:hyparam} (right), the model attains the highest mAPs on known label proportions 10\%, 50\%, 90\%, when $s=0.6$.

\begin{table}
    \parbox{0.32\linewidth}{
        \centering
        \scriptsize
        \setlength{\tabcolsep}{2pt}
        \caption{Ablation studies of LogicMix on MS-COCO. In mAP\%.}
        \vspace{6pt}
        \label{tab:logicmix_ablation}
        \begin{tabular}{cccccccc}
            \hline
            MLE & MMS & A$\rightleftharpoons$O & 10\% & 50\% & 90\% & Avg. \\
            \hline
            \checkmark & \checkmark & \checkmark & \textbf{76.3} & \textbf{83.5} & \textbf{85.1} & \textbf{81.6} \\
            \checkmark &            & \checkmark & 75.9 & 83.2 & 85.0 & 81.4 \\
            \checkmark & \checkmark &            & 69.4 & 82.0 & 83.7 & 78.4\\
            \checkmark &            &            & 71.8 & 82.3 & 84.1 & 79.4 \\
                       & \checkmark & \checkmark & 74.0 & 82.4 & 84.9 & 80.5 \\
                       &            & \checkmark & 73.0 & 82.0 & 84.8 & 80.0 \\
                       & \checkmark &            & 67.9 & 78.8 & 83.1 & 76.6 \\
                       &            &            & 68.3 & 79.2 & 83.5 & 77.0 \\
            \hline
            
        \end{tabular}
    }
    \hfill
    \parbox{0.66\linewidth}{
        \centering
        \includegraphics[width=0.66\linewidth]{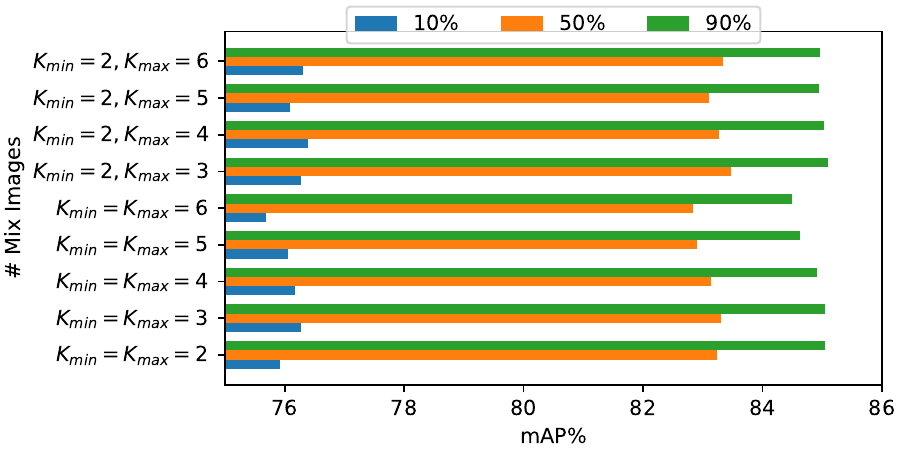}
        \includegraphics[width=0.33\linewidth]{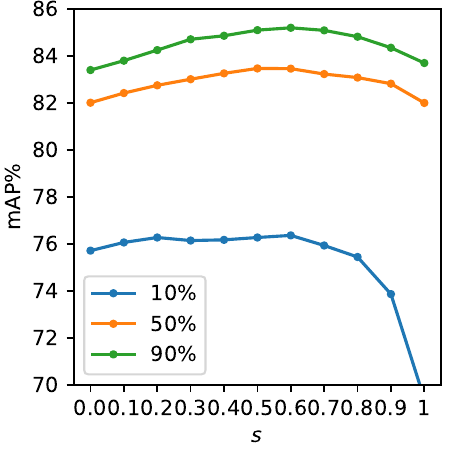}
        \vspace{-6pt}
        \captionof{figure}{Impact of LogicMix's hyperparameters to mAP\% on MS-COCO. (Numerial results in Appendix \ref{sec:hyparam_num})}
        \label{fig:hyparam}
    }
    
    \vspace{-20pt}
\end{table}

\section{Conclusion}
\label{sec:conclusion}
This paper proposes a Mixup variant that supports mixing multiple partially labeled samples, called LogicMix, to bring Mixup in MLC-PL tasks. Comprehensive experiments on three datasets show its remarkable performance and generality across different benchmarking scenarios. Moreover, we demonstrate its outstanding collaboration abilities with recent advanced methods and achieve new SOTA performances. To sum up, LogicMix is promising to be a general data augmentation method of great importance in MLC-PL.
\paragraph{Limitations} The experiment setting implicitly assumes labels are clean. The effectiveness of LogicMix under noisy, partially labeled settings has yet to be investigated. Besides, its effectiveness on specific domains of images (\eg, medical images) is to be studied.
\paragraph{Broader Impacts} The contributions help address the challenge of using less training data to build accurate MLC models. It might enable individuals/small companies with fewer resources to enjoy the deep learning techniques and might help break the monopoly of resource-rich enterprises on AI.

\begin{ack}
This work is supported by Macao Polytechnic University under grant number RP/FCA-04/2022.
\end{ack}

\bibliography{main}

\bibliographystyle{abbrvnat}


\appendix

\section{Appendix / supplemental material}

\subsection{Visual Illustration of the Framework with LogicMix}
\label{sec:framework_visual}

\begin{figure}[H]
    \centering
    \includegraphics[width=\linewidth]{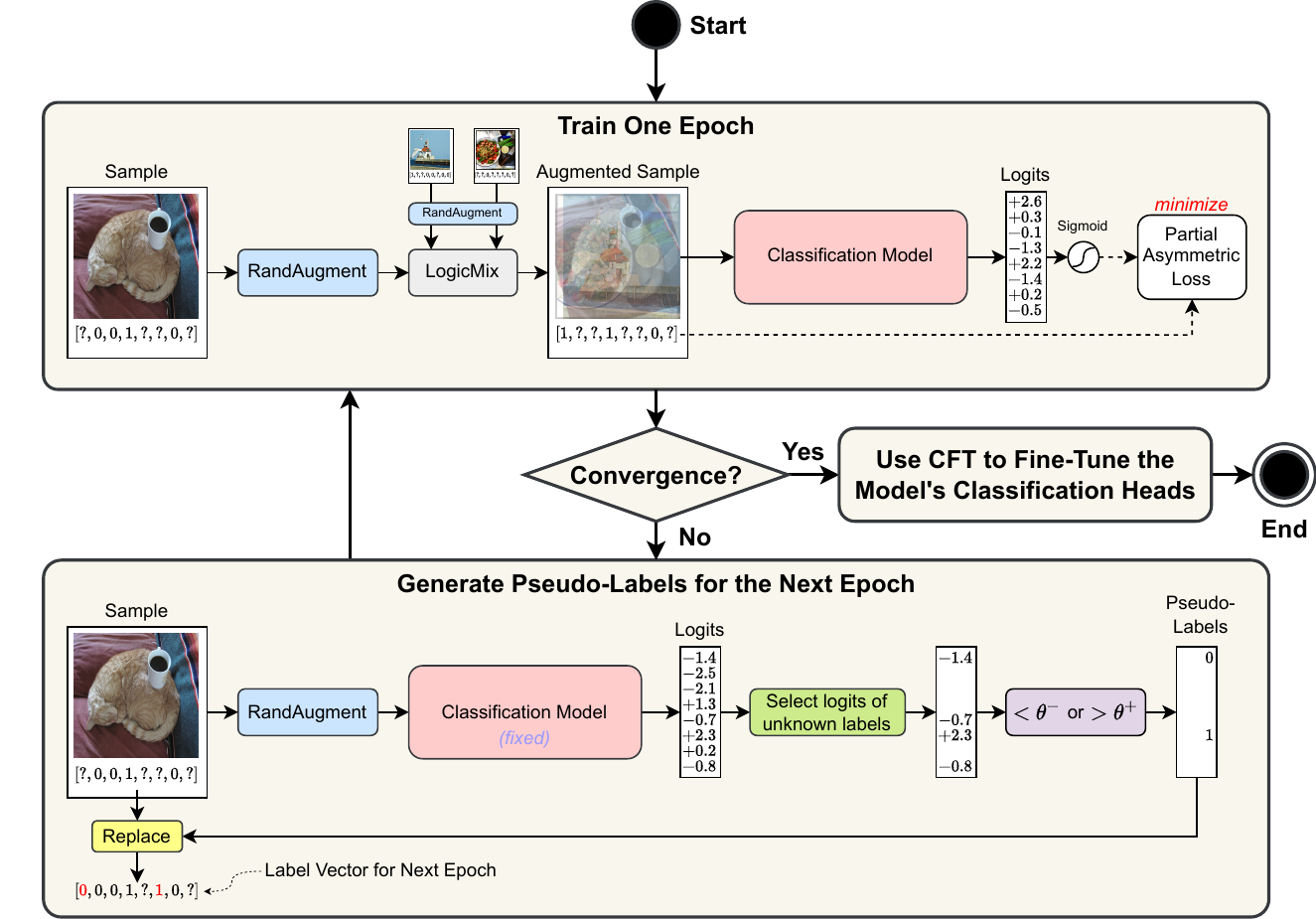}
    \caption{The overview of the framework with LogicMix.}
    \label{fig:framework}
\end{figure}

\subsection{More Details of the Experimental Setups}
\label{sec:experimental_setups_detail}
\paragraph{Datasets} 
\textbf{MS-COCO} \cite{lin_microsoft_2014} consists of 80 categories. The training and test sets consist of 82,783 and 40,502 image samples, respectively.
\textbf{VG-200} \cite{chen_structured_2022} is derived from Visual Genome \cite{krishna_visual_2017}. The latter consists of 108,249 images with labels of 80,138 categories. VG-200 is obtained by keeping the top 200 most frequent categories and randomly splitting the images into a training set (98,249 images) and a test set (10,000 images).
\textbf{Pascal VOC 2007} \cite{everingham_pascal_2009} consists of 20 categories. The training and test sets consist of 5,011 and 4,952 images, respectively.

\paragraph{Compute Resources} Each major experiment is run on a RTX 4090 24G GPU (around 10 hours each). Each auxiliary experiment is run on either a RTX A6000 48G GPU or a Tesla V100 32G GPU. All experiments are capable to be run on a single RTX 4090 24G GPU.

\subsection{Collaboration of LogicMix with SST, HST, and LL-R}
\label{sec:cooperation_others}
Here, we demonstrate the collaboration of LogicMix to SST \cite{chen_structured_2022}, SARB \cite{pu_semantic-aware_2022}, and LL-R \cite{kim_large_2022}.
For the experiments on SST and SARB, we add LogicMix ($p=0.1, K_{min}=K_{max}=2$) to the official codes provided in the original papers. Both SST and SARB frameworks originally used cropping to augment samples. We insert LogicMix after the cropping, meaning LogicMix combines samples that have been cropped. We run the modified official codes without changing any hyperparameter.
For LL-R, since the original paper did not provide the codes for partially labeled datasets (only codes for single positive labels is available). We re-implement LL-R ($\Delta_{rel}=1$) with the same implementation details in Section \ref{sec:implementation_details}.

The results are shown in \autoref{tab:logicmix_cooperation}. LogicMix can improve mAP in most cases. Note that the results of SST/SARB + LogicMix may be further improved by properly adjust the original hyperparameters.

\FloatBarrier
\begin{table}[h!]
    \centering
    \footnotesize
    \caption{Results of collaboration of LogicMix to SST, SARB, and LL-R on MS-COCO.}
    \label{tab:logicmix_cooperation}
    \vspace{6pt}
    \begin{tabular}{lcccc}
        \hline
        Method & 10\% & 50\% & 90\% \\
        \hline
        SST \cite{chen_structured_2022} & 68.39 & 78.28 & 80.07 \\
        SST + LogicMix & \textbf{69.19} & \textbf{78.61} & \textbf{80.29} \\
        \hline
        SARB \cite{pu_semantic-aware_2022} & 68.89 & 78.29 & \textbf{80.36} \\
        SARB + LogicMix & \textbf{68.97} & \textbf{78.52} & 80.00 \\
        \hline
        LL-R \cite{kim_large_2022} & 70.51 & 82.27 & 84.44 \\
        LL-R + LogicMix & \textbf{72.74} & \textbf{83.28} & \textbf{85.21} \\
        \hline
    \end{tabular}
\end{table}
\FloatBarrier

\subsection{Results of Framework with Other Model Architectures}
\label{sec:framework_architecture}

\FloatBarrier
\begin{table}[h!]
    \centering
    \footnotesize
    \caption{Results of the framework with different model architectures on MS-COCO.}
    \label{tab:framework_architecture}
    \vspace{6pt}
    \begin{tabular}{clcccc}
        \hline
        Model Architecture (Pretraining) & Method & 10\% & 50\% & 90\% \\
        \hline
        \multirow{2}{*}{ResNet-101 \cite{he_deep_2016} (ImageNet-1K)}
        & RandAugment+CL+CFT & 78.4 & 84.2 & 85.3 \\
        & RandAugment+CL+CFT + LogicMix & \textbf{79.0} & \textbf{85.0} & \textbf{86.2} \\
        \hline
        \multirow{2}{*}{TResNet-M \cite{ridnik_tresnet_2021} (ImageNet-21K)}
        & RandAugment+CL+CFT & 80.3 & 85.0 & 86.2 \\
        & RandAugment+CL+CFT + LogicMix & \textbf{80.6} & \textbf{85.7} & \textbf{86.8} \\
        \hline
        \multirow{2}{*}{\makecell{SwinTransformerV2-Base ($384^2$) \cite{liu_swin_2022}\\(ImageNet-22K $\rightarrow$ ImageNet-1K)*}}
        & RandAugment+CL+CFT & 84.2 & 88.2 & 89.2 \\
        & RandAugment+CL+CFT + LogicMix & \textbf{85.7} & \textbf{89.3} & \textbf{90.2} \\
        \hline
        
    \end{tabular}

    \raggedright
    ~~~* Global batch size 128 = batch size 8 $\times$ gradient accumulation of 16 batches.
\end{table}
\FloatBarrier

\subsection{Implementation Details of the Other Mixup Variants}
\label{sec:implementation_details_other_mixup}

\paragraph{Mixup \cite{zhang_mixup_2017} (AN)}
\begin{align}
    \mathbf{x}' &= \lambda \mathbf{x}^{(1)} + \lambda \mathbf{x}^{(2)} \\
    y'_c &= \lambda \mathtt{f}(y_c^{(1)})+ (1-\lambda) \mathtt{f}(y_c^{(2)}), \text{ for }c=\{1, 2, \dots, C\} \\
    & \text{where } \lambda \sim Beta(\alpha, \alpha)
\end{align}
where $\mathtt{f}(\cdot)$ is to treat an unknown label as a negative label
\begin{equation}
    \mathtt{f}(y) = \begin{cases}
        y &, y\in\{0, 1\}\\
        0 &, y=?
    \end{cases} 
\end{equation}.
We follow the original paper \cite{zhang_mixup_2017} to set $\alpha=0.2$.

\paragraph{Wang \etal \cite{wang_baseline_2019} (AN)}
\begin{align}
    \mathbf{x}' &= \lambda \mathbf{x}^{(1)} + (1 - \lambda) \mathbf{x}^{(2)} \\
    y'_c &= \mathtt{f}(y_c^{(1)}) \lor \mathtt{f}(y_c^{(2)}),  \text{ for }c=\{1, 2, \dots, C\} \\
    & \text{where } \lambda = 0.5
\end{align}.
This Mixup is enabled/disabled alternately for every epoch.

\paragraph{Multi-label Mixup \cite{gao_dynamic_2022} (AN)}
\begin{align}
    \mathbf{x}' &= \lambda \mathbf{x}^{(1)} + (1 - \lambda) \mathbf{x}^{(2)} \\
    y'_c &= \mathtt{f}(y_c^{(1)}) \lor \mathtt{f}(y_c^{(2)}),  \text{ for }c=\{1, 2, \dots, C\} \\
    & \text{where } \lambda \sim \mathcal{U}_{[0, 1]} \text{.}
\end{align}

\paragraph{Mixup-PME \cite{dong_revisiting_2022}}
\begin{align}
    \mathbf{x}' &= \lambda \mathbf{x}^{(1)} + \lambda \mathbf{x}^{(2)} \\
    y'_c &= \begin{cases}
        ? &, y_c^{(1)} = ? \\
        \lambda \hat{\mathtt{f}}(y_c^{(1)})+ (1-\lambda) \hat{\mathtt{f}}(y_c^{(2)}) &, \text{else}
    \end{cases}, \text{ for }c=\{1, 2, \dots, C\} \\
    & \text{where } \lambda \sim \mathcal{U}_{[\alpha, 1]} 
\end{align}
where $\hat{\mathtt{f}}(\cdot)$ treats an unknown label as $0.5$
\begin{equation}
    \hat{\mathtt{f}}(y) = \begin{cases}
        y &, y\in\{0, 1\}\\
        0.5 &, y=?
    \end{cases} 
\end{equation}.
We follow the original paper \cite{dong_revisiting_2022} to set $\alpha=0.75$.

\subsection{Impact of LogicMix to Training Time}
\label{sec:computation_time}

\begin{figure}[h!]
    \centering
    \includegraphics[width=0.3\linewidth]{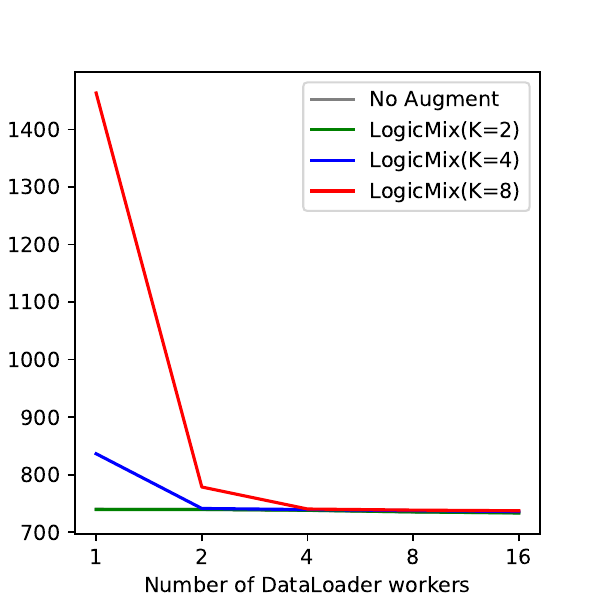}
    \caption{The mean computation times per training epoch of using LogicMix with different number of data loader workers.}
    \label{fig:computation_time}
\end{figure}

In this section, we report the model training time of using LogicMix with different numbers of dataloader workers.
Experiments are conducted on a consumer-grade desktop computer with Intel i9-13900KF CPU, 128 GB RAM, and Ubuntu 22.04.1 LTS. Model are trained on MS-COCO dataset using a RTX 4090 24G GPU. The dataset is stored in a Samsung 980 Pro 2TB SSD. We use the PyTorch DataLoader (\url{https://pytorch.org/docs/stable/data.html#torch.utils.data.DataLoader}) with different number of workers to preprocess the data. We run each setting for 10 times and report the mean times with standard deviation.
\autoref{fig:computation_time} shows the results and \autoref{tab:computation_time} reports the numerical results. As can be seen, when using the number of workers $\geq 4$, the computation times of using LogicMix are reduced to the same level as using no augmentation. It implies that increasing the number of workers successfully minimizes LogicMix's impact on the training time to be negligible.

\FloatBarrier
\begin{table}[h!]
    \centering
    \footnotesize
    \caption{The mean computation times per training epoch (in second, with standard deviations) of using LogicMix with different number of data loader workers.}
    \label{tab:computation_time}
    \vspace{6pt}
    \begin{tabular}{ccccc}
    \hline
        \# Workers & No augment & LogicMix($K=2$) & LogicMix($K=4$) & LogicMix($K=8$)\\
        \hline
        1 & 738.71 $\pm$0.29 & 739.65	$\pm$0.18& 836.44	$\pm$16.30&1463.14	$\pm$34.84\\
        2 & 738.87 $\pm$0.14 & 739.65	$\pm$0.20&741.37	$\pm$0.16&778.65	$\pm$14.57\\
        4 & 738.27	$\pm$0.50 & 738.23	$\pm$0.56&739.45	$\pm$0.42&740.41	$\pm$0.37\\
        8 & 735.55	$\pm$0.34 & 735.36	$\pm$0.33&737.59	$\pm$0.41&738.51	$\pm$0.40\\
        16 & 733.41	$\pm$0.20 & 733.48	$\pm$0.28&735.96	$\pm$0.31&737.41	$\pm$0.48\\
        \hline
    \end{tabular}
\end{table}
\FloatBarrier

\subsection{Numerical Results of the Hyperparameters Analysis on LogicMix}
\label{sec:hyparam_num}

\FloatBarrier
\begin{table}[h!]
    \centering
    \scriptsize
    \caption{Numerical results of the hyperparameters analysis on LogicMix on MS-COCO. In mAP\%.}
    \vspace{6pt}
    \begin{tabular}{ccccccc}
        \hline
        $K_{min}$ & $K_{max}$ & $s$ & 10\% & 50\% & 90\% & Avg.\\
        \hline
        2 & 2 & 0.5 & 75.9 & 83.2 & 85.0 & 81.4 \\
        3 & 3 & 0.5 & 76.3 & 83.3 & 85.0 & 81.5 \\
        4 & 4 & 0.5 & 76.2 & 83.1 & 84.9 & 81.4 \\
        5 & 5 & 0.5 & 76.0 & 82.9 & 84.6 & 81.2 \\
        6 & 6 & 0.5 & 75.7 & 82.8 & 84.5 & 81.0 \\
        2 & 3 & 0.5 & 76.3 & 83.5 & 85.1 & 81.6 \\
        2 & 4 & 0.5 & 76.4 & 83.3 & 85.0 & 81.6 \\
        2 & 5 & 0.5 & 76.1 & 83.1 & 84.9 & 81.4 \\
        2 & 6 & 0.5 & 76.3 & 83.3 & 85.0 & 81.5 \\
        \hline
        2 & 3 & 0.0 & 75.7 & 82.0 & 83.4 & 80.4 \\
        2 & 3 & 0.1 & 76.1 & 82.4 & 83.8 & 80.8 \\
        2 & 3 & 0.2 & 76.3 & 82.7 & 84.3 & 81.1 \\
        2 & 3 & 0.3 & 76.1 & 83.0 & 84.7 & 81.3 \\
        2 & 3 & 0.4 & 76.2 & 83.3 & 84.9 & 81.4 \\
        2 & 3 & 0.5 & 76.3 & 83.5 & 85.1 & 81.6 \\
        2 & 3 & 0.6 & 76.4 & 83.5 & 85.2 & 81.7 \\
        2 & 3 & 0.7 & 75.9 & 83.2 & 85.1 & 81.4 \\
        2 & 3 & 0.8 & 75.4 & 83.1 & 84.8 & 81.1 \\
        2 & 3 & 0.9 & 73.9 & 82.8 & 84.4 & 80.3 \\
        2 & 3 & 1.0 & 69.4 & 82.0 & 83.7 & 78.4 \\
        \hline
    \end{tabular}
    \label{tab:my_label}
\end{table}
\FloatBarrier

\subsection{Licenses for Used Assets}
\label{sec:licenses}
\begin{itemize}
    \item MS-COCO (2014) dataset \cite{lin_microsoft_2014}. Multiple licenses \footnote{\url{https://cocodataset.org/\#termsofuse}}.
    \item VG dataset \cite{krishna_visual_2017}. CC BY 4.0.
    \item VG-200 \cite{chen_structured_2022}\footnote{\url{https://github.com/HCPLab-SYSU/HCP-MLR-PL}}. Unknown license.
    \item Pascal VOC 2007 \cite{everingham_pascal_2009}. Unknown license.
    \item PyTorch \cite{ansel_pytorch_2024}. Custom license \footnote{\url{https://github.com/pytorch/pytorch?tab=License-1-ov-file\#readme}}.
    \item timm \footnote{\url{https://github.com/huggingface/pytorch-image-models}}. Apache-2.0 license.
    \item Part of code from \cite{ben-baruch_multi-label_2022} \footnote{\url{https://github.com/alibaba-miil/partiallabelingcsl}}. MIT License.
    \item Part of code from \cite{chen_structured_2022}\footnote{\url{https://github.com/HCPLab-SYSU/HCP-MLR-PL}}. Unknown license.
    
\end{itemize}

\end{document}